\documentclass[letterpaper, 10 pt, conference]{ieeeconf}  
\usepackage[bookmarks=true]{hyperref}
\usepackage{algorithm}
\usepackage{algpseudocode}
\usepackage{graphicx}
\usepackage{float}
\usepackage{multirow}
\usepackage{array}
\usepackage[table]{xcolor}
\usepackage{amsmath}

\IEEEoverridecommandlockouts                              

\title{\LARGE \bf
Force Feedback Control For Dexterous Robotic Hands Using Conditional Postural Synergies}

\author{
Dimitrios Dimou\textsuperscript{1} \and 
Jos\'e Santos-Victor\textsuperscript{1} \and 
Plinio Moreno\textsuperscript{1} 
\thanks{
\textsuperscript{1}Institute for Systems and Robotics, 
Instituto Superior Tecnico, Universidade de Lisboa, Portugal.
Emails: {\tt\small mijuomij@gmail.com, \{jasv, plinio\}@isr.tecnico.ulisboa.pt}}%
}

\begin{document}

\maketitle
\thispagestyle{empty}
\pagestyle{empty}

\begin{abstract}

We present a force feedback controller for a dexterous robotic hand
equipped with force sensors on its fingertips. Our controller uses 
the conditional postural synergies framework to generate the grasp 
postures, i.e. the finger configuration of the robot, at each time step
based on forces measured on the robot's fingertips. Using this framework
we are able to control the hand during different grasp types using only 
one variable, the grasp size, which we define as the distance between 
the tip of the thumb and the index finger. Instead of controlling the 
finger limbs independently, our controller generates control signals 
for all the hand joints in a (low-dimensional) shared space (i.e. synergy space).
In addition, our approach is modular, which allows to execute various
types of precision grips, by changing the synergy space according 
to the type of grasp. We show that our controller is able to lift 
objects of various weights and materials, adjust the grasp 
configuration during changes in the object's weight, and perform 
object placements and object handovers. 
\end{abstract}


\section{Introduction}
To perform complex manipulation tasks in unstructured environments,
humans use tactile feedback from their fingers. This feedback is
provided by tactile afferents located in the skin of the hand. Particularly, for
handling small objects with precise movements, the afferents located in the
fingertips are used, which have high density and adapt fast to pressure changes
\cite{Johansson2009CodingAU}. These afferents provide information about the
characteristics of the exerted contact forces, such as the magnitude and the
direction. 
For anthropomorphic robots to be able to perform 
dexterous tasks similar force feedback signals must
be used to alleviate problems arising from
uncertainty in measurements, and handle external 
perturbations. For example, using open-loop position 
control to lift a heavy object may fail due to slip
without any feedback mechanism to provide tactile 
information.

Previous works have used tactile sensors to design
force controllers that use slip prediction to update
the desired normal forces applied by the fingertips.
The slip predictors are based on machine learning 
models such as neural networks and random forests 
to classify multi-modal signals from a tactile sensor.
In all previous works, each finger was separately
controlled by an independent force controller. In addition,  
they required labeled data to train the slip predictors 
and because each finger is controlled independently 
is not obvious how to implement different anthropomorphic 
grasp types.

In this work we develop a force controller that takes as input the force
readings of the fingertips and computes the grasp size which is then used along
with a grasp type label to generate a grasp posture with the desired
characteristics. To avoid slippage the desired normal contact force is
calculated to be proportional to the tangential contact forces.  The applied
normal force is then controlled using the size of the grasp as a control
variable. Larger grasp sizes mean less force is applied to the object.  So the
grasp size is calculated from the error between the desired normal force and the
actual measured normal force. The grasp size is then given to the posture
sampler that generates a grasp posture, i.e. the finger joint angles. The
posture sampler is modeled with a conditional Variational Auto-Encoder (cVAE)
based on the framework proposed in \cite{9560818}. With this framework 
we abstract away the low-level control of the fingers and generate hand postures
based on high-level properties such as the type and the size of the grasp.
So it works as a mapping function that takes as input a low-dimensional vector
and the grasp type and size as conditional variables and maps them to a set 
of joint angles.


We show that with our controller we can control a dexterous 
robotic hand to lift objects of different weights using three 
precision grasps. Our controller is also able to compensate 
and retain a stable grasp during changes in the objects' weight,
for example when filling up a cup or emptying it. In addition 
we show how with the addition of the hand pose information we 
can use the controller to calculate if the tangential force 
is due to gravity or due to a support surface and use this 
information to perform handovers and place down objects on 
surfaces. We perform several real-world experiments with a 
dexterous robotic hand to showcase the capabilities of our 
controller and support our design choices. 

To sum up our main contributions are 
\begin{itemize}
    \item We develop a controller for a dexterous robotic hand
    that uses force feedback and the conditional synergies
    framework to perform dexterous manipulation tasks.
    
    \item We show that with our controller we can easily use
    different precision grasp types, by changing only the 
    grasp type variable which is given to the grasp posture
    mapping function.
    
    \item We demonstrate by incorporating information about 
    the world pose of the hand we can use our controller to 
    perform additional tasks such as placing down and 
    handing over objects. 
    
\end{itemize}


\section{Related Work}
Roboticists have looked for inspiration in humans for developing
methods for complex object manipulation \cite{Yousef2011TactileSF}. 
Neuroscientists have studied for a long time the processes that
allow humans to use tactile feedback to perform complex manipulation
tasks. Humans tend to adjust the grip force according to the object's
weight, its friction and they use a safety margin to account for
uncertainties \cite{Westling2004FactorsIT}. To gather information 
about the tactile states they use multiple afferents that are 
located in the skin of the fingers \cite{Johansson2009CodingAU}.
There are different afferents in different parts of the hand
depending on their usage, e.g. fast adapting afferents in 
the fingertips for precise manipulation. Based on signals
from these afferents, humans encode simple contact events 
into action phases, such as grasping, lifting or releasing, 
which they combine in order to perform more complex and 
long-horizon manipulation tasks \cite{Flanagan2006ControlSI}.

In robotics tactile sensors have been used for object 
stabilization and slip prediction in a variety of settings.
For example, in \cite{ComplTac}, a compliant anthropomorphic
prosthetic hand was controlled using force sensing to
maintain object stability and avoid slip. In \cite{IntegratedFT},
they develop a control approach that uses integrated force and 
spatial tactile signals to avoid slip with unknown objects
in real world settings. In \cite{TactileMetric1,TactileMetric2},
grasp quality metrics are computed based on the tactile 
feedback from the robots fingertips. In these works,
simple two or three fingered grippers were considered
for simple grasping tasks. 

Force control with anthropomorphic robotic hands has also 
been explored in more recent works.
In \cite{Su2015ForceEA}, they employ three slip prediction 
methods to estimate when slip starts and based on the force
signals at that moment they calculate the friction coefficient
value. Based on the calculated friction coefficient, they 
design a force controller that independently controls each
finger to achieve a desired normal force. The desired normal 
contact force is set to be proportional to the tangential 
contact force and a safety margin based on the evidence 
found in \cite{Westling2004FactorsIT}.
In \cite{Veiga2018InHandOS}, they train a random forest 
to classify the contact states into the classes: 
no contact, contact, slip. Based on this classification
signal, when slip is detected they increase the desired 
normal contact force to avoid it.
In \cite{Deng2020GraspingFC} they train a recurrent 
neural network to estimate slip and the object material
from the readings of a Biotac sensor. The force controller
is increasing the desired normal contact force when slip
is detected.
All these works \cite{Deng2020GraspingFC,Veiga2018InHandOS,Su2015ForceEA}
use tactile feedback sensors to predict slip. 
They collect labeled data, on which they train
their models. This approach is based on complex
and expensive tactile sensors, and the process 
of collecting data is cumbersome. In addition, 
the data do not cover all possible hand poses, 
which would be impractical.

In contrast, in our work we do not rely on
slip prediction, we avoid slip by defining
a tangential force gain and  a safety margin
that work for a large number of objects. 
Furthermore, instead of independently 
controlling each finger we use a synergistic
framework to generate grasp postures, that is
conditioned on two variables: the grasp type and 
the grasp size. This way, instead of controlling 
the values of each joint of each finger, we control
only the two conditional variables greatly 
simplifying the control pipeline. This also,
gives us the ability to use different grasp
types in our manipulation tasks by changing 
only the grasp type variable. 
In \cite{Kent2017RoboticHA}
also a synergistic framework was used to prevent an object
from slipping from a humanoid hand, but they modeled only 
one synergy for a tripod grasp and they used the forces on
the robotic arm as feedback, while we use force feedback from
the fingertips.
Our control algorithm could also be applied to different hands
as it does not depend on the hands configuration.
Finally, in previous approaches only lifting tasks
had been considered. In our work we demonstrate 
that our approach can be used to perform more complex 
tasks, such as placing objects on surfaces and performing 
handovers, which was not done in previous works.

\begin{figure}[h] \centering
\includegraphics[width=0.27\textwidth]{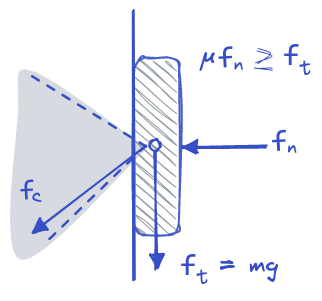}
\caption{\label{fig:contact_example} 
Example of modeling the contacts and friction during manipulation. }
\end{figure}


\section{Methods}
Our goal in this work is to design a control algorithm
for an anthropomorphic robotic hand to perform dexterous
manipulation skills such as lifting and placing down objects.
Our control algorithm will use tactile feedback
from the force sensors on the fingertips of the hand to 
decide the forces that need to be applied to the object
in each step of the task. Given the desired forces to 
be applied, the size of the grasp will be computed.
Given the grasp size and a desired grasp type, 
the posture generator will generate a grasp posture,
i.e. the hand configuration, such that the force constraints
are satisfied.

To model the contacts and friction we use Coulombs' law, which states that
in order to avoid slip, the normal contact force $f_n$ to the
contact surface of an object, times the fiction coefficient $\mu$,
has to be larger than the tangential force $f_t$ \cite{RoboticsHandbook}: 
$$ \mu f_n \geq f_t $$
You can see an example in Figure \ref{fig:contact_example},
where an object is pressed against a wall by an applied 
normal force $f_n$, and we have the tangential 
force $f_t = mg$ due to gravity. In order for the object 
to remain stable we need to apply a normal force:
$$f_n \geq \frac{f_t}{\mu} $$
where $\mu$ is the friction coefficient between the object and the wall.
In the case of a dexterous hand manipulating an object, we want 
the normal forces applied by all fingers to be greater than
the tangential force divided by the friction coefficient of the 
materials of the object and the fingertip. 

\begin{figure*}[t]
\centering
\includegraphics[width=0.8\textwidth]{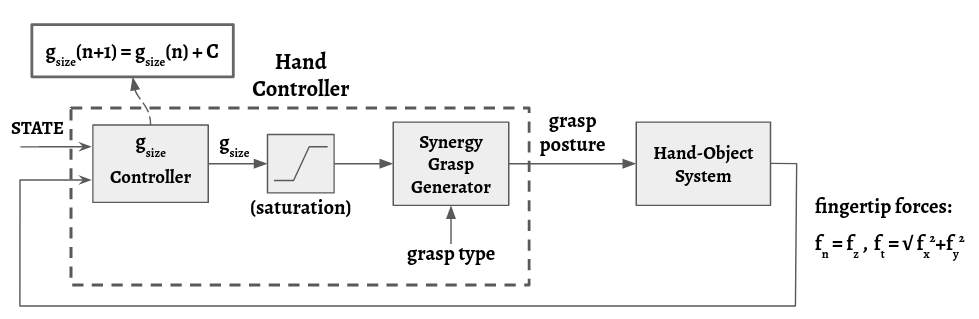}
\caption{\label{fig:force_controller} 
Schematic representation of the proposed force controller. 
The input is the state (GRASP or RELEASE) and the force readings. 
Based on that the grasp size is adjusted by a value $C$ and is given
to the posture mapping function along with the desired grasp type. 
A finger configuration is then generated and commanded to the robot.}
\end{figure*}

Since it is hard to accurately compute the friction coefficient 
between all possible object materials previous works have 
used multi-modal tactile sensors like the BioTac sensor,
which provides information about the pressure, skin deformation, 
and temperature, to predict slip and based on that signal to 
increase the applied normal force. In our work we use the FTS3 sensors \cite{fts3}
which is a low-cost sensor that measures the 3D force applied in each fingertip.
In addition, previous works gathered labeled datasets 
in order to train their slip prediction models which is 
time-consuming and limits the possible orientations of the 
hand, because gathering labeled data for all possible orientations
is impractical. To overcome this we experimentally selected the
parameters that determine the value of the applied normal force
such that we avoid slip for all objects in our dataset, 
from the lightest to the heaviest.

In order to guarantee contact between the fingertip 
and the object, in the beginning of the grasping phase,
we use an offset $f_n^{offset}$ as the minimum
normal force applied by each finger. In \cite{Westling2004FactorsIT} 
they also suggest that humans use an additional safety margin 
which is proportional to the tangential force, $f_n^{margin} \propto f_t$. 
So the final desired normal contact force becomes:

$$ f_n^{des} = G \cdot f_t + f_n^{offset}$$

where $G$ is the gain that includes the friction coefficient
and the additional safety margin.

To alleviate the effects of noise in the sensors, the 
running average of the measured normal force $f_n$ 
and tangential force $f_t$ is used, as a
low pass filter. So for each force measurement we 
have the following relation:

$$ \hat{f}(n+1) = \alpha f(n+1) + (1 - \alpha) f(n) $$
where $\alpha \in (0, 1)$ is a parameter that determines how
much new measurements affect the value, 
and is experimentally selected.

\begin{figure}[h]
\includegraphics[width=0.45\textwidth]{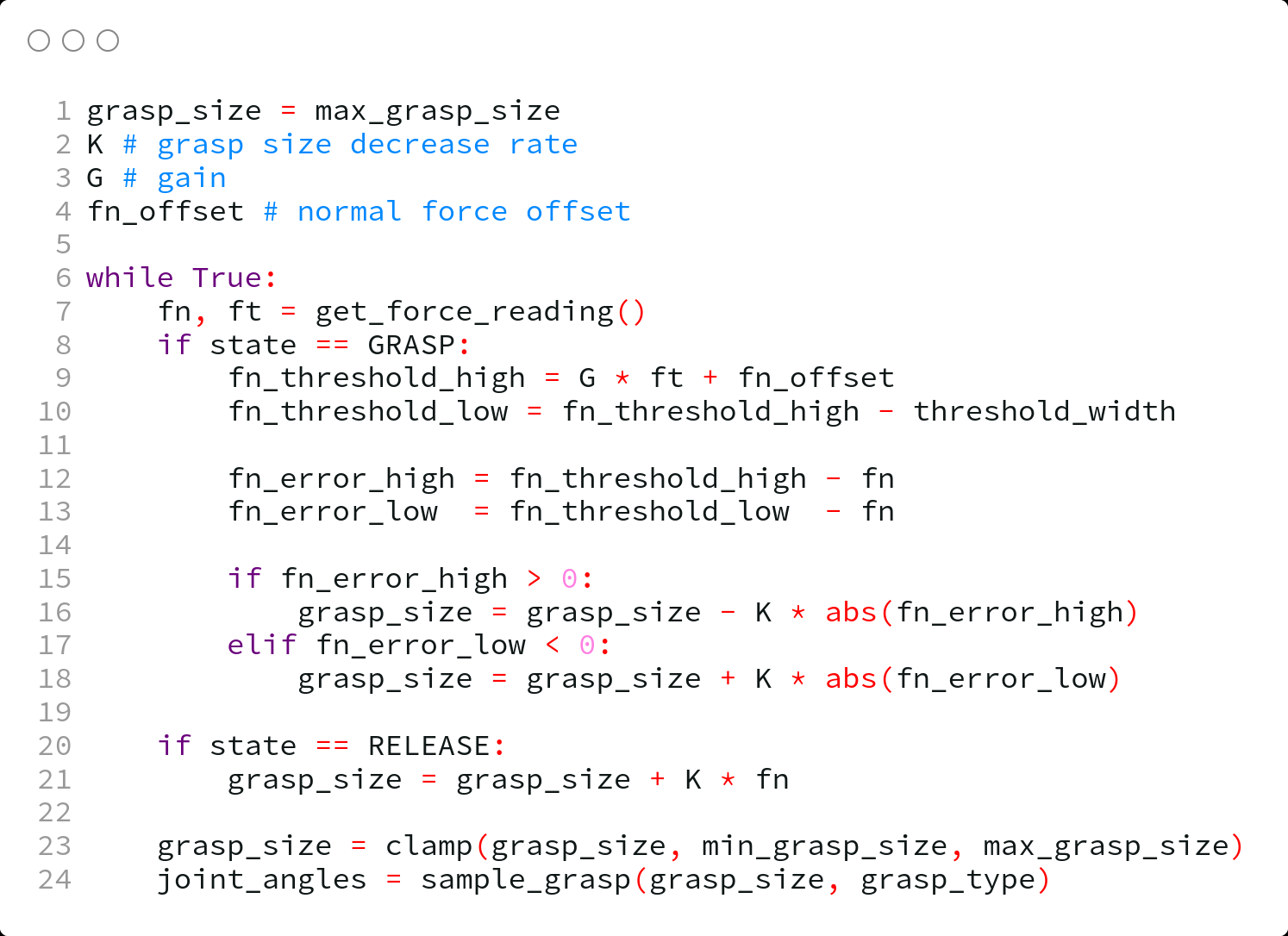}
\caption{\label{fig:algorithm} Our control algorithm in Python-like pseudocode. }
\end{figure}

Given the measured normal force $f_n$ from the fingertip sensors
we can compute the error $f_n^{err}=f_n^{des}-f_n$. 
We use this error signal to control the grasp size
variable $g_{size}$, that we use as a conditional 
variable in our posture mapping function. The grasp size 
represents the distance between the thumb and the 
index finger in a grasp posture. So a smaller grasp
size will result in a tighter grasp and greater normal
force applied to the surface of the object. 
We use a linear controller for the grasp size variable
that is implemented as follows:

$$ g_{size}(n+1) = g_{size}(n) - K \cdot f_n^{err}$$

where $K$ is a parameter that controls the rate
of decrease of the grasp size, and is experimentally selected.
So when the error between the desired normal force
and the actual normal force is large the grasp size 
decreases so tighter grasp postures are generated
in order to apply more normal force.
In practice, in order to avoid oscillations in 
the grasp size we use the desired normal force 
as a \textit{high} threshold that we want the measured 
normal force to be below:
$$ f_n < f_n^{des} = f_n^{threshold\_high}.$$
If the normal force is below that threshold the grasp size does not
change even if there are small oscillations in 
the measured tangential and normal forces.
Also, in order to avoid the hand applying too 
much force that damages the hardware or the object
we use a \textit{low} threshold, that is: 
$$f_n > f_n^{threshold\_low} = f_n^{threshold\_high} - w_{threshold},$$
where $w_{threshold}$ is the width of the threshold in $mN$.
If the measured normal force is below the grasp size increases 
in order to apply less force. So the final grasp size 
variable for grasping is calculated as follows:

\renewcommand{\theequation}{\arabic{equation}}
\begin{equation}\label{eqn:1}
g_{size} (n + 1) = \begin{cases}
g_{size}(n) - K \cdot f_n^{err\_high} &\text{if $f_n^{err\_high} > 0$} \\
g_{size}(n) + K \cdot f_n^{err\_low} &\text{if $f_n^{err\_low } < 0$}
\end{cases}
\end{equation}

\begin{equation*}
\text{where} \begin{cases}
f_n^{err\_high} = f_n^{threshold\_high}- f_n \\
f_n^{err\_low} = f_n^{threshold\_low} - f_n
\end{cases}
\end{equation*}

This is similar to the \textit{deadband} control method \cite{deadband}, 
where instead of having a fixed reference point, an operating range
is set. If the response is in this range, the controller does not
exert any correction. In our case, the operating range changes 
according to the force signals from the robot's fingertips.

The grasp posture mapping function is based on the conditional
postural synergies model presented in \cite{9560818}. 
It uses a conditional Variational Auto-Encoder model
to generate grasps postures conditioned on additional variables
such as the grasp size. In this work we augment this 
model to also generate grasp postures conditioned
on the grasp type. The model is trained on a set of 
labeled grasp samples acquired by teleoperating a 
robotic hand using a data-glove. Using this model
we are able to abstract away the low-level control
of each joint of each finger and generate grasps
based on more general characteristics such as
the type and the size of the grasp. In this way
we can control all the fingers jointly by a single 
value, the grasp size, thus greatly reducing 
the control parameters. In addition we are able 
to use the same control algorithm for different
precision grasp types, by changing the grasp type
conditional variable.

Finally, we can modify our controller to release 
objects instead of grasping them. Given the pose 
of the hand in the world coordinate frame, which 
we can acquire from the robotic arm that is 
attached to, we can use the forward kinematics of
the hand to compute the poses of each fingertip. 
Then using the force readings of each fingertip we
can calculate the global direction of the net 
tangential force. If the angle between the direction
of the net tangential force and the direction of 
gravity is less than $90$ degrees, i.e. the net 
tangential force's direction is towards the ground,
we assume that the tangential force is due to gravity
pulling the object, so the force controller tries 
to grasp it. If the angle is more than $90$ degrees, 
i.e. the net tangential force's direction is upward, 
it means that something is pushing (or pulling) the 
object upward, in which case we assume that the object
is touching on a support surface or someone is pulling
the object so the controller increases the grasp size
given to the posture mapping function proportionally to the 
normal force measured thus slowly releasing the object. 
Opening the grasp is done by controlling the grasp 
size variable as follows:

\begin{equation}\label{eq:2}
g_{size} (n + 1) = g_{size}(n) + K \cdot f_n
\end{equation}

That way we can place objects on surfaces but also perform
robot to human handovers, where the robot holds the object
and the human grasps the object and slightly pushes or pulls it up,
signaling to the robot that there is a support surface. 
The robot then slowly releases the object by opening its 
grasp. We showcase these scenarios in the experiments' section.

Based on these observations, we present our force 
controller in Figure \ref{fig:force_controller}. The hand 
starts in an open pre-grasp position, a latent point is 
sampled from the prior distribution of the posture mapping function, and
given the desired grasp type and the grasp size
a grasp posture, i.e. the joint angles of the fingers,
is sampled. The initial grasp size is set to the maximum value,
and when the force controller comes into effect and depending
on the state of the system and the forces on the fingertips
grasp size changes by some value $C$, according 
to equations  \ref{eqn:1},\ref{eq:2}, until the desired 
normal force is achieved.

To choose between grasping or releasing an object we 
use a finite state machine formulation. When the hand
reaches the desired grasp pose, which we assume is 
provided, the \textit{GRASP} state is activated, in 
which the controller tries to grasp the object. When 
the controller detects that the tangential force 
applied to the object is coming from a support surface
the state changes to the \textit{RELEASE} state, in
which the controller releases the object by opening 
the grasp. You can see the full algorithm in 
Python-like pseudocode in Figure \ref{fig:algorithm}.

To summarize, the advantages of our controller 
compared with previous approaches are threefold:
1) instead of controlling each joint of each 
finger of the hand we use only two variables, 
the grasp size and the grasp type, which allows 
us to perform multiple grasp types by changing 
only one variable while the grasp size variable 
is common among all grasp types, that greatly 
reduces the complexity of the control process 
compared to independently controlling a 21 DoF 
hand to perform different grasp types, 
2) we do not rely on slip prediction for 
controlling the desired normal force, which 
involves gathering labeled data and works only 
for the hand poses in the training dataset, and 
3) we can use our controller to also release 
objects instead of only grasping them.

\begin{figure*}[t]
\centering
\includegraphics[width=0.93\textwidth]{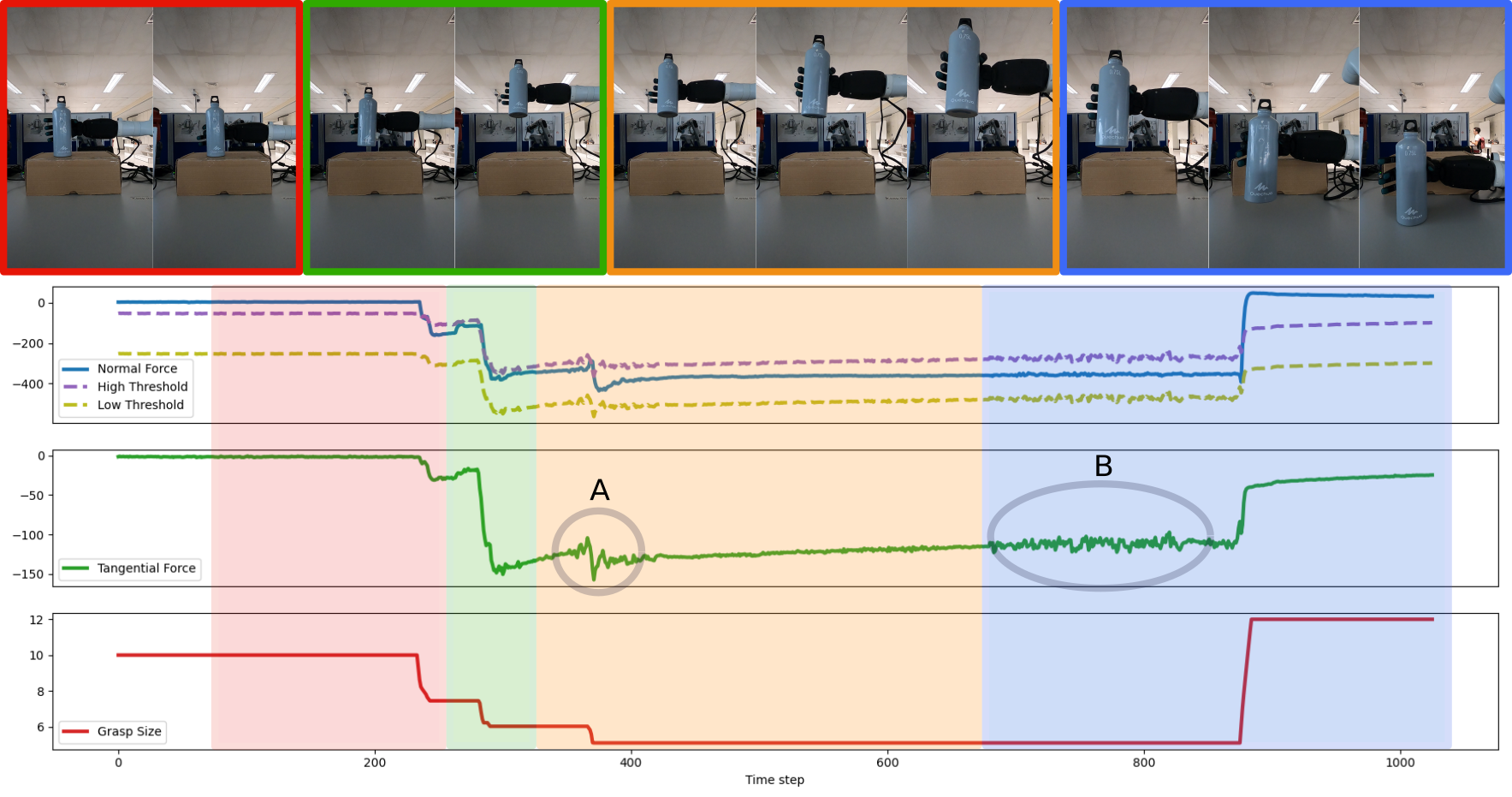}
\caption{\label{fig:pick_and_place} 
Our first experiment. The robot picks up a bottle, transports it, and places down on the desk.
In the bottom part of the figure, you can see the control signals during this task.}
\end{figure*}


\section{Experimental Results}
\subsection{Experimental Set-up.}

\begin{figure}[h]
\centering
\includegraphics[width=0.32\textwidth, height=0.22\textwidth]{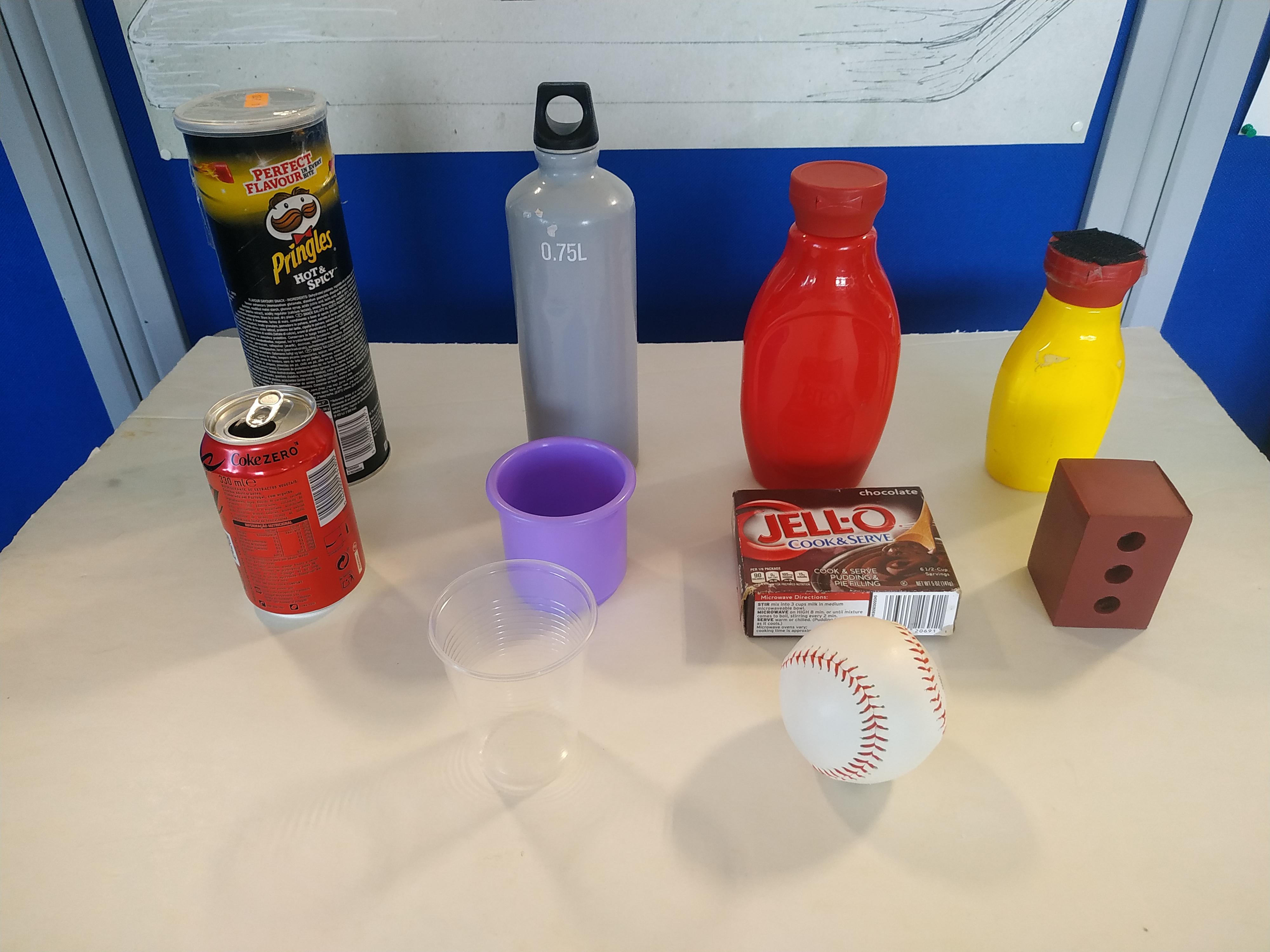}
\caption{\label{fig:objects} The household objects used in our experiments.}
\end{figure}

For our experiments we used the Seed Robotics 
RH8D Hand \cite{robot}, which is a robotic hand
with 7 DoFs. The hand is equipped with the FTS-3
force sensors \cite{fts3} in each fingertip, which
are high resolution tactile sensors that provide 
the 3D force applied in each fingertip. 
The sensor provides data at a rate of 50Hz. 
For the experiments the hand was mounted on a Kinova Gen3 7DoF robot. 
To train the posture mapping function we used the CyberGlove to 
teleoperate the hand and collect 468 grasps belonging 
to three precision grasp types: tripod, pinch, lateral 
tripod. The architecture of the cVAE model was the same as in
\cite{9560818}, with the addition of the grasp type
as a conditional variable, which was one-hot encoded.
We used 10 household objects shown in Figure \ref{fig:objects}. 
With the heaviest object weighing $380g$ and the lightest $1g$.
During the experiments the trajectories of the arm were
prerecorded, while the hand was controlled online by our 
control algorithm.

\begin{figure*}[t]
\centering
\includegraphics[width=0.93\textwidth]{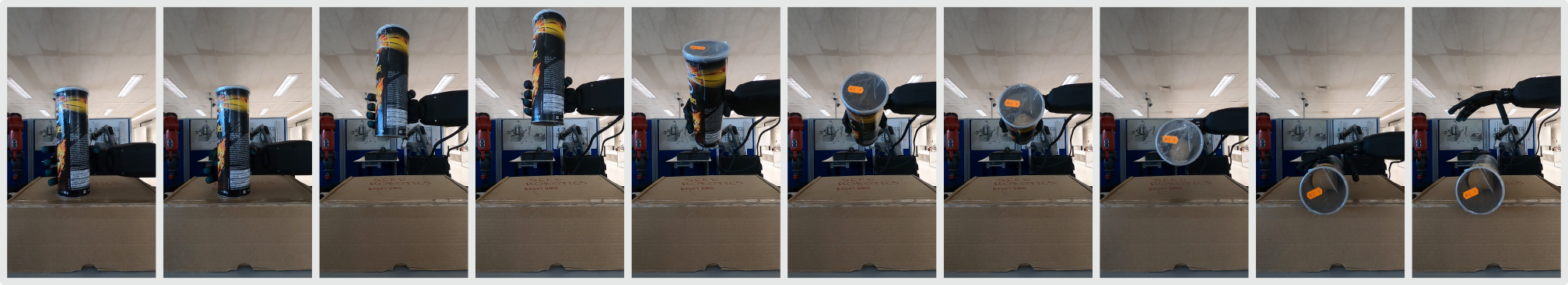}
\includegraphics[width=0.93\textwidth]{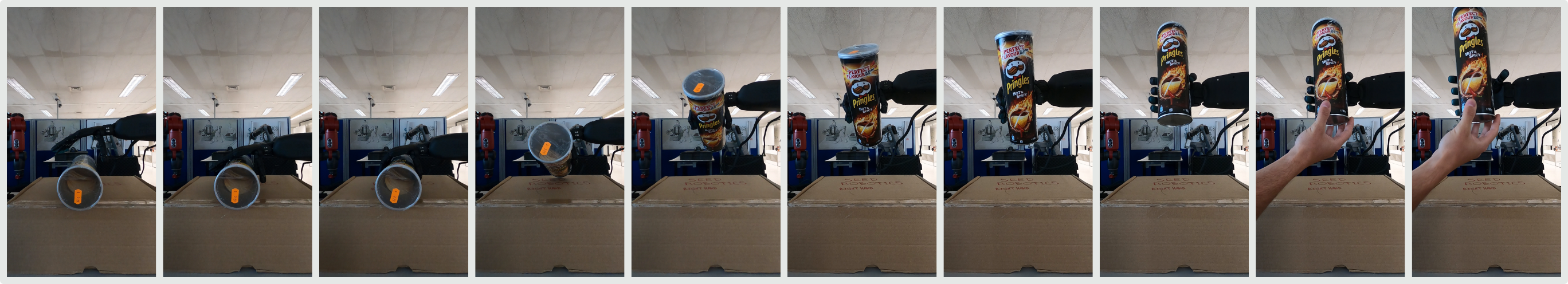}
\includegraphics[width=0.93\textwidth]{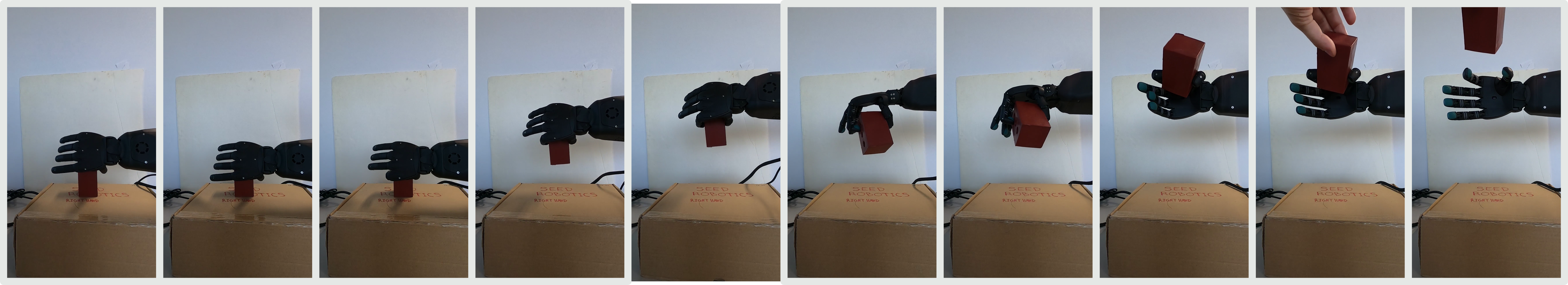}
\caption{\label{fig:other_experiments} 
In the upper row of images, you can see our second experiment. The robot picks up
the chips can, rotates it 90 degrees, and places back down. In the middle row, 
for our third experiment, the robot picks up the chips can, rotates it 90 degrees, 
and hands it over to a person. In the bottom row, for our forth experiment,
the robot picks up a foam brick, rotates it 180 degrees, and hands it over to a 
person, using a pinch grasp.}
\end{figure*}

\begin{figure*}[t!]
\centering
\includegraphics[width=0.93\textwidth]{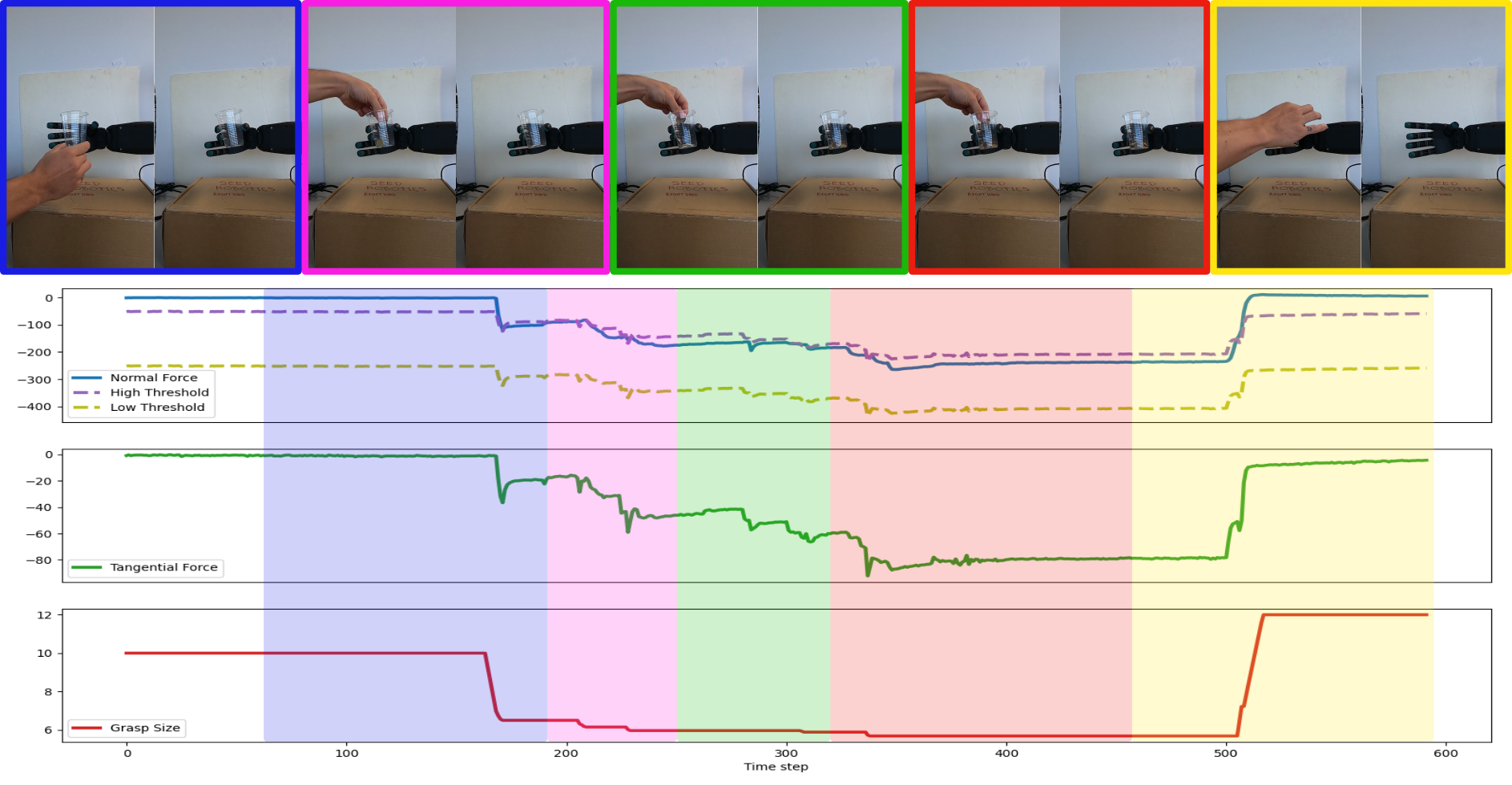}
\caption{\label{fig:cup_add_weight} 
In our fifth experiment, a person hands over an empty plastic cup to the robot, 
throws coins in it to increase its weight while the robot adjusts its grip 
to stabilize the object, and then hand overs the cup back to the person.}
\end{figure*}

\subsection{Parameter tuning.}
To select the values of the parameters in our
controllers we conducted preliminary experiments
where we tested lifting and releasing several 
objects, with different physical properties. 
To select the value of the normal offset force $f_n^{offset}$,
we used an empty plastic cup as our test object, 
and we choose a value such that the fingers do not 
deform the cup. The final value of the parameter was
set to -50 mN.
To select the values of the gain $G$ and the rate 
of decrease $K$, of the grasp size, we experimented 
with the heaviest object in our dataset, which is 
the mustard bottle and weighs $380g$. The gain $G$ 
was set to $2.0$ such that the desired normal force
would be enough to hold the object. The rate of 
change of the grasp size was set to $100.0$, based 
on the operating frequency of the force sensor and 
the range of values of the tangential force. 
For the tangential force averaging process we used 
a parameter value of $\alpha_t = 0.7$, because we 
want the controller to be sensitive to fast changes in its
value, that can arise for example during lifting an object. 
For the normal force averaging process we used a parameter
value of $\alpha_n = 0.5$, as we do not want it to be 
affected by noise that could make the controller overconfident.


\subsection{Experiments.}
To explore the capabilities of our controller, 
we demonstrate five experiments of increasing complexity:
1) we picked and placed a bottle using a tripod grasp,
2) we picked, rotated and placed a chips can on a box using a tripod grasp,
3) we picked, rotated and handed over the chips can to a person using a tripod grasp,
4) we picked, rotated and handed over a brown foam brick to a person using a pinch grasp,
5) a person handed over a plastic cup to the robot, filled it with coins to 
increase its weight, and the robot then handed it back to the person using a tripod grasp.

You can see the execution of the first experiment in Figure \ref{fig:pick_and_place}.
Under the pictures of the execution you can see the signals
recorded by the controller: the average normal force applied by 
all fingers (blue line), the thresholds $f_n^{threshold\_high}.$ 
(purple dashed line) and $f_n^{threshold\_low}.$ (yellow dashed line),
the average tangential force (green), and the grasp size used in 
each time-step (red). The task is divided four stages: 1) (red part) 
the initial grasp of the object, in this stage the force 
controller closes the grasp until the applied normal force is below 
the offset $f_n^{offset}$, 2) (green part) the robot lifts the object, 
as it tries to lift the tangential force increases, 
increasing the threshold, so the grasp size decreases 
to apply more normal force, 3) (orange part) the robot transports
the object, you can see, in point \textit{A} in the Figure, a 
perturbation in the tangential force when the robot begins to move, 
the controller responds by decreasing the grasp thus stabilizing the
object, and 4) (blue part) the robot enters the releasing phase, 
where it lowers the arm until it detects that the tangential force
is due to a support surface, then it stops lowering the 
arm and increases the grasp size slowly releasing the object.
In point \textit{B} in the Figure, you can see that there is
noise in the tangential force, due to the arm moving to place the 
object on the table, that is also reflected in the desired 
normal force. Because we use the desired normal force as a 
threshold and not as a reference signal this noise is not  
manifested in the control of the grasp size. 
You can see the execution of the second experiment in 
the upper part of Figure \ref{fig:other_experiments}.
This experiment demonstrates the ability of the controller
to handle arbitrary hand poses. The experiment is divided
in four parts: 1) the robot enters the \textit{GRASP} phase
and the force controller generates grasps to achieve a 
normal contact force below the $f_n^{offset}$ threshold, 
2) the robot lifts the object and adjusts the grasp size
to avoid the object falling, 3) the hand rotates to place
the chips can on the horizontal position, and 
4) the robot enters the \textit{RELEASE} phase, and the arm
lowers until the object touches the box, when the hand 
detects the supporting surface, it starts to slowly release
the object.
You can see the execution of the third experiment 
in the middle part of Figure \ref{fig:other_experiments}.
This experiment demonstrates the ability of the controller
to perform robot to human handovers. The experiment is divided
in four parts: 1) the robot enters the \textit{GRASP} phase
and the force controller generates grasps to achieve a 
normal contact force below the $f_n^{offset}$ threshold, 
2) the robot lifts the object and adjusts the grasp size
to avoid the object falling, 3) the hand rotates to place
the chips can on the vertical position, and 
4) the robot enters the \textit{RELEASE} phase, 
the arm stays still, the human grasps the object from the 
bottom and slightly pushes it up, the hand then detects that
there is a supporting surface and starts to slowly release 
the object.
You can see the execution of the fourth experiment
in the bottom part of Figure \ref{fig:other_experiments}.
This experiment is similar to previous one, but the grasp 
type that the robot uses is a pinch grasp, that involves 
only the thumb and the index finger. To perform this we only
had to alter the grasp type conditional variable that was 
given to the posture mapping function.
You can see the execution of the fifth experiment in the
bottom part of Figure \ref{fig:cup_add_weight}. In the first part (blue)
of the experiment the robot closes its grasp, by reducing the grasp
size, until the normal force is below the force offset. In the next
three parts (pink, green, red) the person throws coins in the cup
to increase its weight. You can see in the signal plots that each
time coins are added the tangential force decreases so the normal
force threshold decreases too. The grasp sizes then decreases as
well in order to apply more normal force.
This experiment demonstrates the ability of the controller to
handle perturbations in the weight of the object during grasping.


\section{Conclusion}
In summary, we presented a controller that uses
force feedback integrated with conditional synergies
to control a dexterous robotic hand to grasp and release
objects. We demonstrated that our controller can
lift objects of different weights and materials 
while avoiding slip, react online when the weight 
of the object changes, place them down on surfaces, 
and hand them over to humans. In addition, the control 
architecture is modular, so the synergy grasp mapping
component can be easily changed in order to control
several precision grasp types. However, 
our experiments also revealed various 
limitations of our controller. For example 
our method fails to stabilize the object when
rotational slip occurs. In addition hardware limitations
such as, slow update rates and noise in the force 
measurements can create problems that result in 
the object falling. In future work we plan to
incorporate additional sensing modalities, such as
vision to alleviate some of these issues.

\section*{Acknowledgment}
Work  partially supported by 
the H2020 FET-Open project \textit{Reconstructing the Past: Artificial
Intelligence and Robotics Meet Cultural Heritage} (RePAIR) under EU grant
agreement 964854, and by the Lisbon Ellis Unit (LUMLIS),
and the FCT PhD grant [PD/BD/09714/2020].


\bibliographystyle{IEEEtran}
\bibliography{root}

\begin{thebibliography}{10}
\providecommand{\url}[1]{#1}
\csname url@samestyle\endcsname
\providecommand{\newblock}{\relax}
\providecommand{\bibinfo}[2]{#2}
\providecommand{\BIBentrySTDinterwordspacing}{\spaceskip=0pt\relax}
\providecommand{\BIBentryALTinterwordstretchfactor}{4}
\providecommand{\BIBentryALTinterwordspacing}{\spaceskip=\fontdimen2\font plus
\BIBentryALTinterwordstretchfactor\fontdimen3\font minus
  \fontdimen4\font\relax}
\providecommand{\BIBforeignlanguage}[2]{{%
\expandafter\ifx\csname l@#1\endcsname\relax
\typeout{** WARNING: IEEEtran.bst: No hyphenation pattern has been}%
\typeout{** loaded for the language `#1'. Using the pattern for}%
\typeout{** the default language instead.}%
\else
\language=\csname l@#1\endcsname
\fi
#2}}
\providecommand{\BIBdecl}{\relax}
\BIBdecl

\bibitem{Johansson2009CodingAU}
R.~S. Johansson and J.~R. Flanagan, ``Coding and use of tactile signals from
  the fingertips in object manipulation tasks,'' \emph{Nature Reviews
  Neuroscience}, vol.~10, pp. 345--359, 2009.

\bibitem{9560818}
D.~Dimou, J.~Santos-Victor, and P.~Moreno, ``Learning conditional postural
  synergies for dexterous hands: A generative approach based on variational
  auto-encoders and conditioned on object size and category,'' in \emph{2021
  IEEE International Conference on Robotics and Automation (ICRA)}, 2021, pp.
  4710--4716.

\bibitem{Yousef2011TactileSF}
H.~Yousef, M.~Boukallel, and K.~Althoefer, ``Tactile sensing for dexterous
  in-hand manipulation in robotics-a review,'' \emph{Sensors and Actuators
  A-physical}, vol. 167, pp. 171--187, 2011.

\bibitem{Westling2004FactorsIT}
G.~Westling and R.~S. Johansson, ``Factors influencing the force control during
  precision grip,'' \emph{Experimental Brain Research}, vol.~53, pp. 277--284,
  2004.

\bibitem{Flanagan2006ControlSI}
J.~R. Flanagan, M.~C. Bowman, and R.~S. Johansson, ``Control strategies in
  object manipulation tasks,'' \emph{Current Opinion in Neurobiology}, vol.~16,
  pp. 650--659, 2006.

\bibitem{ComplTac}
N.~Wettels, A.~R. Parnandi, J.-H. Moon, G.~E. Loeb, and G.~S. Sukhatme, ``Grip
  control using biomimetic tactile sensing systems,'' \emph{IEEE/ASME
  Transactions on Mechatronics}, vol.~14, no.~6, pp. 718--723, 2009.

\bibitem{IntegratedFT}
G.~De~Maria, P.~Falco, C.~Natale, and S.~Pirozzi, ``Integrated force/tactile
  sensing: The enabling technology for slipping detection and avoidance,'' in
  \emph{2015 IEEE International Conference on Robotics and Automation (ICRA)},
  2015, pp. 3883--3889.

\bibitem{TactileMetric1}
R.~Krug, A.~J. Lilienthal, D.~Kragic, and Y.~Bekiroglu, ``Analytic grasp
  success prediction with tactile feedback,'' in \emph{2016 IEEE International
  Conference on Robotics and Automation (ICRA)}, 2016, pp. 165--171.

\bibitem{TactileMetric2}
H.~N. Dang and P.~K. Allen, ``Stable grasping under pose uncertainty using
  tactile feedback,'' \emph{Autonomous Robots}, vol.~36, pp. 309--330, 2014.

\bibitem{Su2015ForceEA}
Z.~Su, K.~Hausman, Y.~Chebotar, A.~Molchanov, G.~E. Loeb, G.~S. Sukhatme, and
  S.~Schaal, ``Force estimation and slip detection/classification for grip
  control using a biomimetic tactile sensor,'' \emph{2015 IEEE-RAS 15th
  International Conference on Humanoid Robots (Humanoids)}, pp. 297--303, 2015.

\bibitem{Veiga2018InHandOS}
F.~Veiga, B.~B. Edin, and J.~Peters, ``In-hand object stabilization by
  independent finger control,'' \emph{ArXiv}, vol. abs/1806.05031, 2018.

\bibitem{Deng2020GraspingFC}
Z.~Deng, Y.~Jonetzko, L.~Zhang, and J.~Zhang, ``Grasping force control of
  multi-fingered robotic hands through tactile sensing for object
  stabilization,'' \emph{Sensors (Basel, Switzerland)}, vol.~20, 2020.

\bibitem{Kent2017RoboticHA}
B.~A. Kent and E.~D. Engeberg, ``Robotic hand acceleration feedback to
  synergistically prevent grasped object slip,'' \emph{IEEE Transactions on
  Robotics}, vol.~33, pp. 492--499, 2017.

\bibitem{RoboticsHandbook}
\BIBentryALTinterwordspacing
B.~Siciliano and O.~Khatib, Eds., \emph{Springer Handbook of Robotics}.\hskip
  1em plus 0.5em minus 0.4em\relax Springer Berlin Heidelberg, 2008. [Online].
  Available: \url{https://doi.org/10.1007%2F978-3-540-30301-5}
\BIBentrySTDinterwordspacing

\bibitem{fts3}
``{Seed Robotics - RH8D Adult Robot Hand},''
  \url{https://www.seedrobotics.com/rh8d-adult-robot-hand}.

\bibitem{deadband}
\BIBentryALTinterwordspacing
H.-H. Ko, J.-S. Kim, J.~Kim, J.-G. Baek, and S.-S. Kim, ``Intelligent adaptive
  process control using dynamic deadband for semiconductor manufacturing,''
  \emph{Expert Systems with Applications}, vol.~38, no.~6, pp. 6759--6767,
  2011. [Online]. Available:
  \url{https://www.sciencedirect.com/science/article/pii/S0957417410014363}
\BIBentrySTDinterwordspacing

\bibitem{robot}
``{Seed Robotics - FTS3 3D High Resolution Tactile (Pressure) sensor},''
  \url{https://kb.seedrobotics.com/doku.php?id=fts:fts3_pressuresensor}.

\end{thebibliography}

\end{document}